\documentclass[runningheads]{llncs}

\usepackage{eccv}

\usepackage{eccvabbrv}

\usepackage{graphicx}
\usepackage{booktabs}

\usepackage{multirow}
\usepackage[accsupp]{axessibility}  
\usepackage{amsfonts}
\usepackage{amsmath}
\usepackage{multirow}
\usepackage{graphicx}
\usepackage{booktabs}
\usepackage{marvosym}

\usepackage{hyperref}

\usepackage{orcidlink}

\begin{document}

\title{2K Retrofit: Entropy-Guided \protect\\ Efficient Sparse Refinement for \protect\\ High-Resolution 3D Geometry Prediction} 

\titlerunning{Abbreviated paper title}

\author{
Tianbao Zhang\textsuperscript{1,3,6\,*} \quad
Zhenyu Liang\textsuperscript{2\,*} \quad
Zhenbo Song\textsuperscript{5} \quad
Nana Wang\textsuperscript{1} \\
Xiaomei Zhang\textsuperscript{5} \quad
Xudong Cai\textsuperscript{5} \quad
Zheng Zhu\textsuperscript{4} \quad
Kejian Wu\textsuperscript{5} \\
Gang Wang\textsuperscript{5} \quad
Zhaoxin Fan\textsuperscript{1\Letter}
}

\authorrunning{F.~Author et al.}
\institute{
$^{1}$~BUAA\quad
$^{2}$~NUDT \quad
$^{3}$~Dim12 AI Inc\quad
$^{4}$~GigaAI \quad
$^{5}$~OpenHelix Robotics \\
$^{6}$~Current Robotics \\
}

\maketitle

\begingroup
\renewcommand\thefootnote{\Letter}
\footnotetext{Corresponding author.}
\endgroup

\begin{center}
    \captionsetup{type=figure}
    \includegraphics[width=1.0\textwidth]{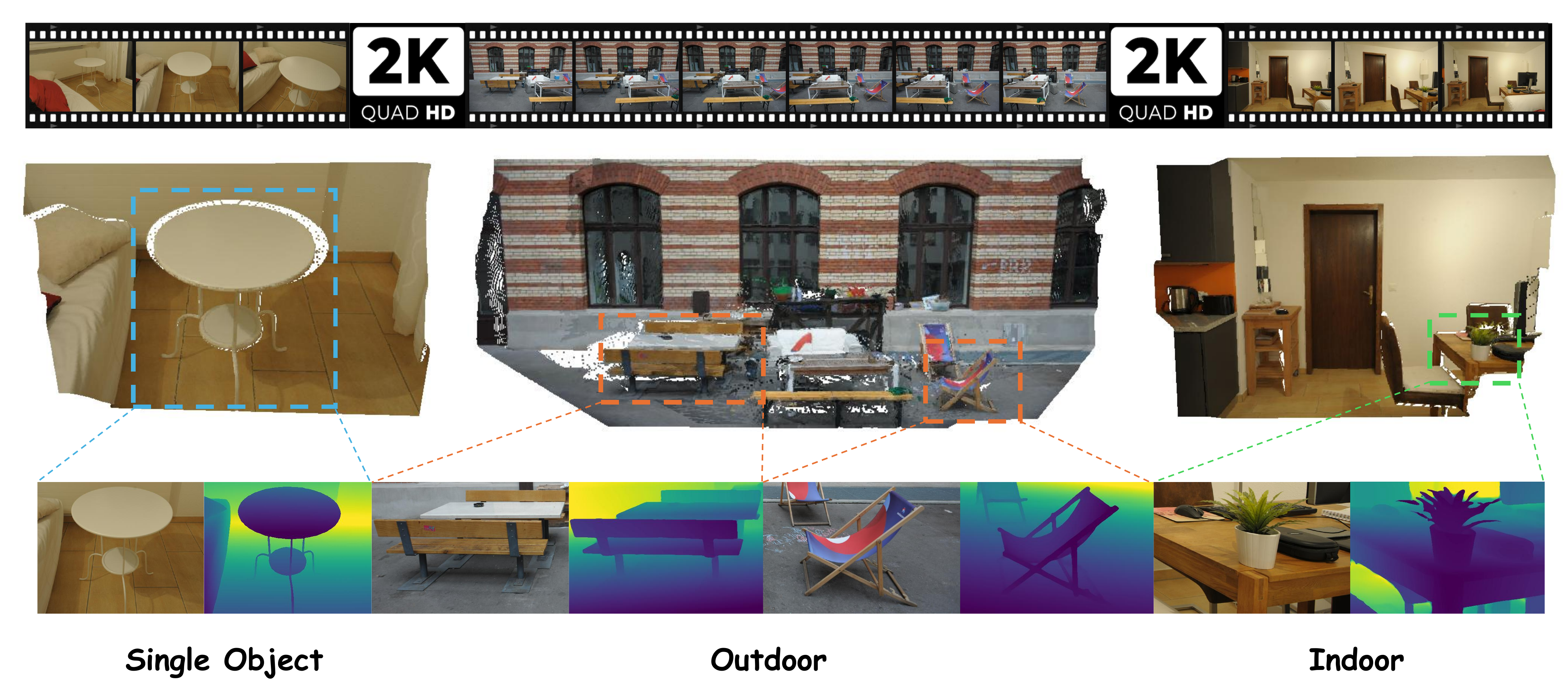}
    \captionof{figure}{\textbf{2K Retrofit} efficiently enhances high-resolution geometric prediction through an entropy-guided sparse refinement strategy. Given 2K-resolution RGB inputs from diverse scenes (single-object, indoor, and outdoor), our framework first performs fast coarse geometric inference using a frozen low-resolution foundation model, then selectively refines only high-uncertainty pixels at full resolution. This design preserves global geometric consistency while recovering fine-scale 3D structures efficiently, achieving accurate and scalable 2K-level depth and pointmap estimation.}
\label{teaser}
\end{center}

\begin{abstract}
    High-resolution geometric prediction is fundamental for robust perception in autonomous driving, robotics, and augmented/mixed reality.
    Existing 3D foundation models demonstrate strong generalization and scalability across diverse real-world scenarios. However, we observe that their capability to handle real-world high-resolution (2K-level) scenes remains constrained by the prohibitive computational and memory requirements of dense inference, making large-scale deployment impractical. To address this limitation, we propose 2K Retrofit—to the best of our knowledge, the first framework enabling existing geometric foundation model to perform
    efficient 2K-resolution inference without backbone modification or retraining. Our method adopts a simple yet effective sparse refinement strategy, where a frozen foundation model provides fast coarse geometric predictions, and an entropy-guided sparse refinement selectively enhances high-uncertainty regions to restore fine-scale 3D structure. This design preserves the global consistency of the backbone while delivering precise, high-fidelity geometry at a fraction of the cost. Extensive experiments on standard benchmarks, including ARKitScenes, ScanNet++, and ETH3D, demonstrate that 2K Retrofit consistently achieves state-of-the-art accuracy and efficiency, bridging the gap between foundation model research and scalable high-resolution 3D vision applications. Code will be released upon acceptance.
\end{abstract}

\section{Introduction}
\label{sec:intro}
High-resolution geometric prediction is fundamental for robust perception and interaction in modern applications such as autonomous driving \cite{zoedepth,2}, embodied intelligence \cite{3,4}, and augmented reality \cite{5}, since fine-grained depth maps and 3D reconstructions serve as the foundation for precise scene understanding, object localization, and manipulation. However, the spatial resolution of depth sensors in practical settings remains a key bottleneck. This inherent resolution imbalance in the captured data leads to the loss of fine-grained geometric cues that are vital for accurate 3D perception. Such resolution disparity thus may severely restrict systems' ability to perceive small-scale structures—such as railings, handles, or cables along corridors—which, from another perspective, are essential for safe navigation and precise interaction. 

\begin{figure}
\centering
\includegraphics[height=4cm,width=0.56\linewidth]{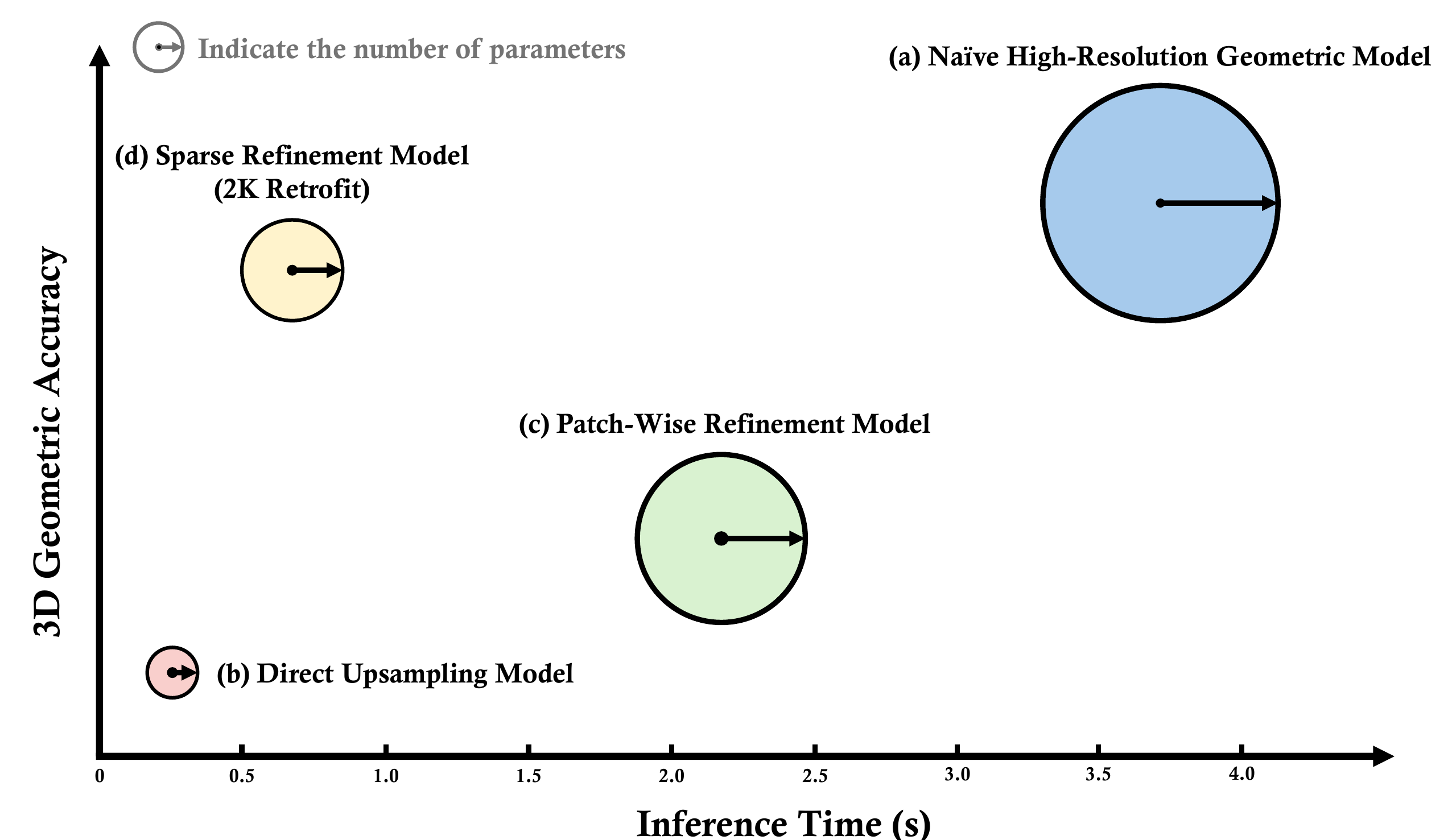}
\caption{\textbf{Differences of 2K Retrofit with exising high-resolution geometry pipelines.}
(a) Native high-resolution geometric models train large networks directly on 2K data, achieving strong fidelity but with high memory and training cost. (b) Direct upsampling models are lightweight and fast but struggle to recover fine 2K structures. (c) Patch-wise refinement models require multiple low-resolution inferences and complex patch fusion, leading to slow inference and spatial inconsistencies. (d) 2K Retrofit (ours) provides efficient inference and competitive 2K geometric fidelity with far fewer parameters.
}\label{fig1}
\end{figure}

Addressing this bottleneck requires 3D geometric model capable of reconstructing high-fidelity 3D structure that lies beyond the native limits of current depth sensors. Although recent approaches have improved geometric prediction through advanced upsampling strategies and model optimization, their inference remains constrained to moderate resolutions (typically around 960×480), leaving true 2K-level prediction an open challenge. Meanwhile, as can be seen, though recent foundation models such as Depth Anything V1/V2 \cite{da,da2}, DUSt3R \cite{dust3r}, VGGT \cite{vggt}, and others \cite{foundationstereo,mvsanywhere,depthpro} trained on large-scale datasets exhibit strong generalization across diverse scenarios. They are typically limited by low-resolution training data and cannot directly deliver precise geometric predictions at high resolution. To overcome this limitation, recent works attempt to imrpove the performance through explicit high-resolution modeling. As illustrated in Fig.~\ref{fig1}, existing approaches fall into three main categories: (1) methods that directly train native high-resolution geometric models~\cite{5,6,7}, aiming to learn fine-grained structure through full-resolution supervision; (2) methods that perform low-resolution inference followed by upsampling or depth super-resolution~\cite{8,bpnet}, leveraging learned priors to enhance spatial detail; and (3) methods that apply patch-wise refinement on top of coarse predictions~\cite{patchfusion, patchrefiner}, focusing computation on local regions to recover high-frequency geometry. However, as discussed, existing methods still fail to achieve efficient and accurate 2K-level 3D geometric prediction that our approach provides.

To address this challenge, we introduce 2K Retrofit—a simple yet effective paradigm that seamlessly augments existing foundation models to produce 2K-resolution geometric outputs, without any modification to the original low-resolution geometric backbone (e.g., Depth Anything, DUSt3R, or VGGT). The central insight underpinning our approach is that the discrepancies between dense low-resolution predictions and their high-resolution counterparts are concentrated in a sparse subset of pixels. By strategically directing refinement only to these high-impact regions, we can unlock precise and reliable high-resolution geometric prediction, while maintaining minimal computational and memory overhead. As shown in Fig.~\ref{teaser}, 2K Retrofit enables efficient high-resolution geometric prediction across diverse scenarios, including single-object, indoor, and outdoor environments.

\vspace{-1em}

\vspace{-2em}

Specifically, 2K Retrofit incorporates an entropy-based pixel selection mechanism that identifies high-uncertainty regions for targeted refinement. These sparse, informative pixels are then processed by a dedicated refinement branch for feature extraction and correction, thereby circumventing the computational burden of full-resolution processing. To facilitate effective training, we further build a large-scale synthetic dataset containing 50,000 images and empirically evaluate our approach on two fundamental 3D vision tasks: depth estimation and 3D reconstruction.
Extensive experiments on ScanNet++~\cite{scannet++}, ARKitScenes~\cite{arkit}, and ETH3D~\cite{ETH} demonstrate that 2K Retrofit achieves state-of-the-art accuracy while maintaining outstanding inference efficiency, surpassing existing methods in both precision and speed. Our contributions can be summarized as follows:

\begin{itemize}
    \item We propose 2K Retrofit, a simple yet effective method that extends 3D foundation models to 2K-resolution geometric prediction via sparse refinement, which matches or surpasses the accuracy of full 2K methods, while offering significantly faster inference.
    \item We introduce sparse geometry refinement, which comprises an entropy-based selector for identifying high-uncertainty pixels, a sparse feature extractor for refinement feature learning, and a gated ensembler for final fusion and refinement.
    \item We conduct extensive evaluations on multiple public high-resolution benchmarks demonstrating the superiority of our approach. Across diverse indoor and outdoor scenarios, our method achieves the best geometric fidelity and more reliably recovers fine-grained 3D structures compared to existing state-of-the-art methods.
\end{itemize}

\section{Related Work}
\label{sec:relatedwork}

\subsection{High-Resolution Monocular Depth Estimation}
Monocular depth estimation seeks to infer scene geometry from a single RGB image, enabling a wide range of 3D vision applications. While deep learning has driven rapid progress, early methods \cite{9,10} were limited by poor generalization across datasets. Later approaches improve robustness via diverse training data \cite{11,12}, affine-invariant losses \cite{13}, and stronger architectures \cite{dpt}. Recently, latent diffusion models \cite{14,marigold} have shown impressive generalization for relative depth, yet are fundamentally limited by input resolution \cite{15,da}, lagging behind the demands of modern high-resolution imaging. To address this gap, Guided Depth Super-Resolution (GDSR) \cite{16,17,18} and tile-based methods \cite{patchfusion,19,20} estimate high-resolution depth by processing image patches independently and stitching results. However, these methods often introduce boundary artifacts and redundant computation. In this paper, we propose a sparse refinement strategy that further improves both accuracy and efficiency.
\begin{figure*}
\centering
\includegraphics[height=6cm,width=1.0\linewidth]{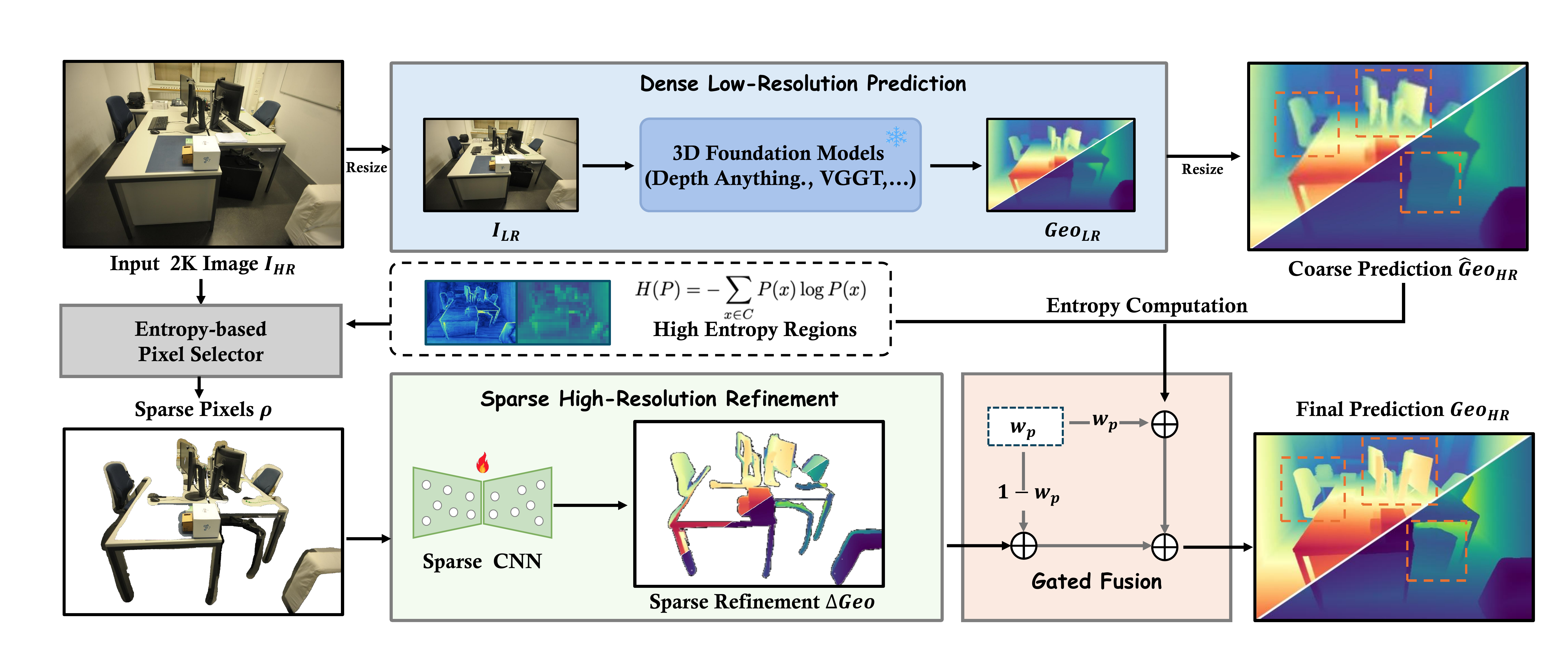}
\caption{\textbf{Framework of 2K Retrofit.} Given an HR input image $I_{\mathrm{HR}}$, we first downsample it and feed the result to a frozen 3D foundation model, upsampling its output to obtain a coarse $\hat{Geo}_{\mathrm{HR}}$. An entropy-based selector then identifies a sparse set of high-uncertainty HR pixels, for which a lightweight sparse extractor predicts localize refinements $\Delta Geo$. A gated ensembler merges the coarse map with $\Delta Geo$, yielding high-fidelity $Geo_{\mathrm{HR}}$ while remaining low-latency, and memory-efficient.}\label{fig2}
\end{figure*}

\subsection{Multi-View Reconstruction}
Multi-View Stereo (MVS) targets dense 3D reconstruction of real-world scenes from multiple overlapping images, providing essential geometric understanding for applications in robotics, mapping, and virtual reality. Conventional MVS pipelines typically rely on known or estimated camera parameters from Structure from Motion (SfM), and can be broadly classified into handcrafted methods \cite{21,22,23}, global optimization strategies \cite{24,25,26}, and deep learning-based techniques \cite{27,28,29,30}. Recent learning-based approaches, such as MVSNet and its variants~\cite{mvsformer,mvsanywhere,mvsformer++},, have significantly advanced reconstruction quality, often requiring depth super-resolution to obtain denser point clouds. Despite this progress, most methods still depend on accurate camera calibration, which can limit their flexibility in unconstrained settings. Emerging feed-forward architectures like DUSt3R \cite{dust3r}, VGGT \cite{vggt}, Dense3R~\cite{dense3r} and Pi3~\cite{pi3} break this dependency, directly predicting dense, aligned point clouds from just a pair of images without explicit camera parameters, thus broadening the applicability of MVS. In contrast to existing high-resolution pipelines, we introduce a model-agnostic sparse refinement framework that seamlessly integrates with modern 3D foundation models to enable efficient and accurate 2K-resolution reconstruction.

\subsection{Sparse Inference in Computer Vision}
Sparse inference—where computation is selectively applied to only a subset of critical activations or parameters—has emerged as an effective strategy for reducing computational cost in computer vision. This selective computation arises naturally in various visual domains, such as videos \cite{31,32}, point clouds \cite{33,34}, and masked images for self-supervised pre-training \cite{35,36,37}. It can also be induced through techniques like activation pruning \cite{38,39,40,41}, token merging \cite{42,43}, or clustering \cite{44}. However, most activation-sparsity approaches are tailored for classification or detection tasks, where maintaining complete spatial structure is less important than capturing global semantics. Beyond feature sparsity, another complementary line of work~\cite{52,53} focuses on spatially selective processing, often referred to as selective refinement, which progressively enhances uncertain or structurally critical regions. In segmentation, for instance, methods such as PointRend~\cite{52} and RefineMask~\cite{53} identify ambiguous pixels at coarse resolutions and selectively refine them using higher-resolution features, significantly improving mask quality. SparseRefine~\cite{sparserefine} is the first work to explicitly introduce the concept of sparse refinement, demonstrating that selectively refining ambiguous regions can significantly improve prediction quality. These approaches have shown strong results in segmentation and other 2D prediction tasks, but they are rarely applied to geometric reasoning. Extending selective refinement to 3D geometry remains underexplored. Inspired by SparseRefine~\cite{sparserefine}, we take a new direction by extending the concept of sparse inference beyond recognition tasks and integrating selective refinement into high-resolution 3D geometry prediction. 

\section{Methodology: 2K Retrofit}
\label{sec:method}

\subsection{Overview}

\begin{figure}
\centering
\includegraphics[height=5cm,width=0.95\linewidth]{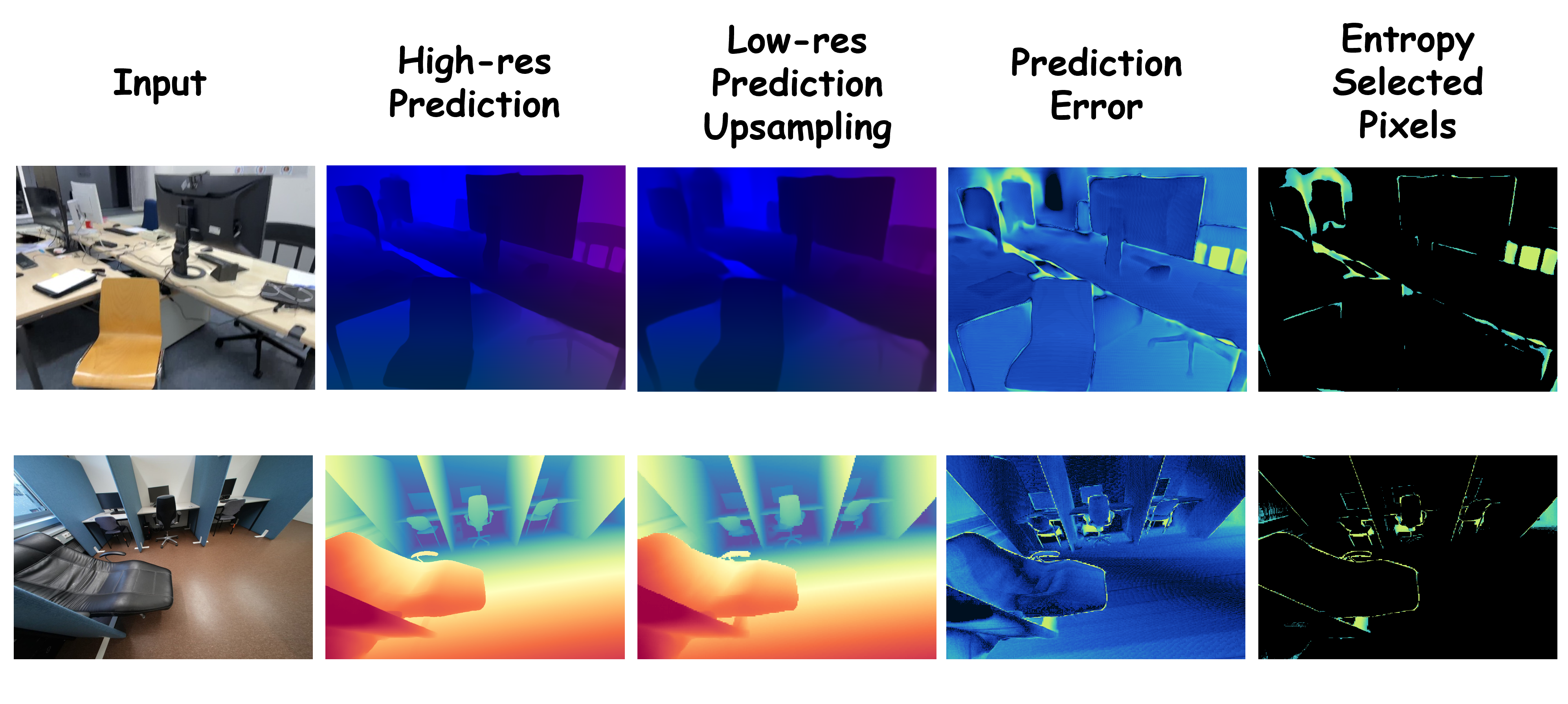}
\caption{Analysis of the correlation between prediction error and entropy-based uncertainty, which confirms a strong correspondence between the sparse pixels that we selected and the error map.} \label{fig3}
\end{figure}

We target high-quality geometric prediction—such as depth estimation and 3D reconstruction—from high-resolution ($2$K) images $\mathbf{I}_{\text{HR}} \in \mathbb{R}^{H \times W \times 3}$. Direct dense inference at this resolution is computationally expensive for existing geometric foundation models. As shown in Fig.~\ref{fig3}, prediction errors at high resolution are typically concentrated on sparse, semantically critical regions (e.g., object boundaries and small structures). This observation motivates a sparse refinement strategy that focuses computation on these regions.

To this end, we propose 2K Retrofit, a framework applicable to both monocular and multi-view settings and compatible with diverse geometric backbones for depth and point map prediction. As illustrated in Fig.~\ref{fig2}, the pipeline consists of two stages: (i) coarse geometric prediction and (ii) sparse high-resolution refinement.

Specifically, the high-resolution image $\mathbf{I}_{\mathrm{HR}}$ is first downsampled to a tractable resolution (e.g., $256$ pixels on the longer side) and processed by a frozen backbone $\mathcal{F}$~\cite{da2,dust3r}. The prediction is then upsampled to $2$K resolution via nearest-neighbor interpolation, producing a dense coarse estimate $\hat{\mathbf{Y}}_{\mathrm{HR}}$.

An entropy-based selector $\mathcal{S}$ computes an uncertainty map and selects high-uncertainty pixels $\mathcal{P}$ (threshold $\alpha$). For each $p \in \mathcal{P}$, a lightweight refinement module $\mathcal{R}$ predicts a local correction from the corresponding high-resolution patch. The correction is fused with the coarse estimate through a gated module $\mathcal{G}$ that adapts weights according to prediction confidence.

The full pipeline—$\mathcal{F}$, $\mathcal{S}$, $\mathcal{R}$, and $\mathcal{G}$—is used in both training and inference. With $\mathcal{F}$ frozen, 2K Retrofit enables 2K inference without retraining and generalizes to monocular or multi-view inputs with different backbones (e.g., Depth Anything~\cite{da2}, VGGT~\cite{vggt}).

\subsection{Dense Low-Resolution Geometric Prediction}
A naïve approach to high-resolution ($2$K resolution in this paper) geometric prediction would be to directly adapt existing foundation models~\cite{da2,vggt} to process $2$K-scale inputs. However, such a strategy is fundamentally constrained by both the immense computational overhead and the limited generalization ability of current models, which are rarely trained or optimized for ultra-high-resolution data. 

To tackle the issue, we adopt a principled two-stage paradigm. Rather than relying on resource-intensive dense prediction at full resolution, we first perform geometric inference on a substantially downsampled version of the input. This low-resolution prediction stage offers high computational efficiency. It also can well leverage the strong global priors learned by foundation models. Hence, the resulting coarse geometric map effectively captures the global structure and semantics of the scene, serving as a robust initialization for subsequent processing.

Specifically, let the high-resolution input image be denoted as $\mathbf{I}_{\mathrm{HR}} \in \mathbb{R}^{H \times W \times 3}$, where $H, W$ indicate the spatial resolution. We first downsample $\mathbf{I}_{\mathrm{HR}}$ to obtain a low-resolution image $\mathbf{I}_{\mathrm{LR}} \in \mathbb{R}^{h \times w \times 3}$, where $h \ll H$ and $w \ll W$. A frozen geometric model $\mathcal{F}$ is then applied to produce a dense low-resolution prediction:

\begin{equation}
    \mathbf{Y}_{\mathrm{LR}} = \mathcal{F}(\mathbf{I}_{\mathrm{LR}}).
\end{equation}
This prediction is then upsampled to the original resolution via nearest-neighbor interpolation, yielding a coarse high-resolution estimate $\hat{\mathbf{Y}}_{\mathrm{HR}} \in \mathbb{R}^{H \times W \times C}$, where $C$ denotes the output dimension (e.g., $C=1$ for monocular depth, $C=3$ for 3D point maps).

\vspace{-2em}
\vspace{-3em}
\subsection{Sparse High-Resolution Geometric Refinement}
While low-resolution predictions deliver substantial gains in computational efficiency, they inevitably struggle to capture fine-grained geometric details, particularly around distant objects, intricate structures, and semantic boundaries. Our systematic error analysis reveals a clear pattern: the vast majority of errors in coarse predictions are not randomly distributed, but instead cluster within sparse, semantically critical regions—such as object contours, thin structures, and areas of category transition (see Fig. \ref{fig3}). These localized regions play an outsized role in scene understanding; thus, selectively improving predictions in these areas can yield significant quality gains with only marginal additional computational cost. 

Building on this observation, we posit that high-fidelity 2K geometric prediction can be achieved by selectively refining only those regions with high error. To this end, we design a targeted refinement approach comprising three core components: an entropy-based selector, a sparse feature extractor, and a gated ensembler. We elaborate on each module in the following sections.
\vspace{-1em}
\subsubsection{(i) Entropy Selector.}
The first step of our refinement pipeline is to identify a sparse subset of pixels that are most in need of correction. Locating these high-error pixels is non-trivial. Fortunately, as shown in Fig.~\ref{fig3}, the regions with the largest prediction errors align closely with high-entropy responses. This strong correlation allows us to use entropy as a principled signal for selecting pixels that truly require refinement. We leverage prediction entropy as a principled measure of model uncertainty, following the intuition that regions with high entropy often correspond to visually or geometrically ambiguous areas such as object boundaries, thin structures, or depth discontinuities. Importantly, since the backbone models are regression models, entropy is \emph{not} computed from the final regression outputs. Instead, we compute entropy from the backbone head features before the final regression projection.

Specifically, based on the coarse prediction, we derive a per-pixel uncertainty map from the backbone logits and apply an entropy-based selector $\mathcal{S}$ to extract a sparse set of high-uncertainty pixels $\mathcal{P} \subseteq \{1,\ldots,H\} \times \{1,\ldots,W\}$. For each pixel $p \in \mathcal{P}$, the entropy is computed from the softmax-normalized logits $\mathbf{q}_p \in \mathbb{R}^C$ as:
\begin{equation}
    \mathcal{H}(p) = -\sum_{c=1}^C \mathbf{q}_p^{(c)} \log \mathbf{q}_p^{(c)}.
\end{equation}
Pixels with entropy above a predefined threshold $\alpha$, i.e., satisfying $\mathcal{H}(p)>\alpha$, are then selected for refinement. This design allows our framework to focus computation on semantically critical yet spatially sparse regions, rather than uniformly refining all pixels. Empirically, our entropy-based selector recovers around 80\% of the high-error pixels while refining only the top 10\% most uncertain ones (Fig.~\ref{fig3}), substantially outperforming magnitude-based selection and approaching the performance of a learnable selector at a fraction of the cost (2.3\,ms vs.\ 4.0\,ms on RTX~4090).
\vspace{-1em}
\subsubsection{(ii) Sparse Refinement Module.}
After identifying the high-uncertainty pixels $\mathcal{P}$, we apply a lightweight sparse refinement module $\mathcal{R}$ to improve local predictions. 
Since the selected pixels are distributed irregularly across the image---similar to sparse 3D point sets with uneven density, occlusion, and complex structure---a dense CNN becomes inefficient and unnecessary. 
Motivated by this analogy, we employ a modified MinkowskiUNet~\cite{45} as our sparse feature extractor, which uses sparse convolutions to operate only on active locations, maintaining sparsity throughout the network and enabling efficient high-resolution processing. 
The refined residual correction for each selected pixel is computed as:
\begin{equation}
    \Delta \mathbf{Y}_p = \mathcal{R}(\mathbf{I}_{\mathrm{HR}}|_{p}), \quad \forall p \in \mathcal{P}.
\end{equation}

\subsubsection{(iii) Gated Fusion.}
After obtaining the sparse refinements, the most straightforward approach is to overwrite the coarse prediction at the refined pixel locations. However, sparse high-resolution predictions lack the global context captured by the dense low-resolution estimates, this naïve substitution often leads to inconsistent or noisy results.
To effectively combine global consistency with local precision, we introduce a gated fusion mechanism that adaptively balances the two sources of information.

To combine the sparse refinements $\Delta \mathbf{Y}$ with the initial dense estimate $\hat{\mathbf{Y}}_{\mathrm{HR}}$, we employ a pixel-wise gating strategy. 
For each selected pixel $p \in \mathcal{P}$, the final prediction is computed as:
\begin{equation}
    \mathbf{Y}_p = f\left(w_p \cdot \hat{\mathbf{Y}}_p + (1 - w_p) \cdot \Delta\mathbf{Y}_p\right),
\end{equation}
where the fusion weight $w_p$ is defined as:
\begin{equation}
    w_p = \sigma\left(\mathrm{MLP}\left([\hat{\mathbf{Y}}_p; \Delta\mathbf{Y}_p; \mathcal{H}(\hat{\mathbf{Y}}_p); \mathcal{H}(\Delta\mathbf{Y}_p)]\right)\right),
\end{equation}
with $\mathrm{MLP}(\cdot)$ a two-layer perceptron and $\sigma(\cdot)$ the sigmoid activation. 

Intuitively, this design allows the network to rely more on the coarse estimate when uncertainty is low, while assigning greater weight to the refined prediction in ambiguous or high-entropy regions. 
By conditioning on both the predictions and their uncertainty, the gated fusion achieves a smooth balance between global structure and local detail, producing coherent and high-fidelity geometric outputs.

\vspace{-1em}
\begin{table*}
    \centering
    \scalebox{0.9}{
    \begin{tabular}{c|c|ccc|ccc}
    \toprule
    \multirow[c]{2}{*}{\raisebox{-0.4ex}{Zero Shot}}& \multirow[c]{2}{*}{\raisebox{-0.4ex}{Method}} &  \multicolumn{3}{c|}{ARKitScenes} & \multicolumn{3}{c}{Scannet++} \\
    \cmidrule{3-8}
    & & AbsRel $\downarrow$ & RMSE $\downarrow$ & $\delta_{0.5}$ $\uparrow$ & AbsRel $\downarrow$ & RMSE $\downarrow$ & $\delta_{0.5}$ $\uparrow$  \\
    \midrule
    \multirow{4}{*}{\raisebox{-0.4ex}{No}} & MSPF~\cite{mspf} & 0.0149 &  0.0363 & 0.9721 & 0.0226 & 0.0975 & 0.9674 \\
    & DepthAnything v2*~\cite{da2} & 0.0326 & 0.0764 & 0.9615 & 0.0371 & 0.1010 & 0.9437 \\
    & PromptDA~\cite{promptda} & 0.0131 & 0.0316 &  0.9825 & 0.0175 & 0.0829 & 0.9781\\
    \cmidrule{2-8}
    & 2K Retrofit (Ours) & \textbf{0.0118} & \textbf{0.0275}  & \textbf{0.9866}  &  \textbf{0.0146} & \textbf{0.0774}  & \textbf{0.9825}  \\
    \midrule
    \multirow{7}{*}{\raisebox{-0.4ex}{Yes}} & Depth Prompting~\cite{DP} & 0.0212 &  0.0422 & 0.9710 & 0.0242 & 0.0983 & 0.9657 \\
    & BPNet~\cite{bpnet} & 0.1261 & 0.2100 & 0.9513 & 0.1419 & 0.1963 & 0.9357 \\
    & DepthAnything v2~\cite{da2} & 0.0411 & 0.0792 & 0.9529 & 0.0402 & 0.1145 & 0.9404 \\
    & Marigold~\cite{marigold} & 0.0609 & 0.1065 & 0.9087 & 0.0603 & 0.1412 & 0.8718 \\
    & PatchRefiner~\cite{patchrefiner} & 0.0206 & 0.0418 & 0.9722 & 0.0237 & 0.0976 & 0.9683\\
    & PromptDA~\cite{promptda} & 0.0142 & 0.0376 & 0.9813 & 0.0224 & 0.0966 & 0.9700 \\
    \cmidrule{2-8}
    & 2K Retrofit (Ours)& \textbf{0.0129} & \textbf{0.0324}  & \textbf{0.9831}  &  \textbf{0.0197} & \textbf{0.0853}  & \textbf{0.9801}  \\
    \bottomrule
    \end{tabular}}
    \caption{Quantitative comparisons on ARKitScenes, Scannet++ dataset. Methods marked with * are finetuned with their released models and code on ARKitScenes~\cite{arkit} and ScanNet++~\cite{scannet++} datasets.}
    \label{tab1}
\vspace{-4em}
\end{table*}

\section{Experiments}
\label{sec:exp}
\subsection{Implementation Details}

\noindent\textbf{Synthetic Training Dataset.}
As the first to address 2K-level geometric prediction, we construct a dedicated synthetic dataset to support our experiments. Inspired by FoundationStereo~\cite{foundationstereo}, we use NVIDIA Omniverse to generate 50K high-quality rendered frames for training both depth estimation and 3D reconstruction models. The dataset covers structured scenes with diverse geometries and textures under complex, realistic lighting. Further dataset details and example scenes are provided in the appendix.
\vspace{-1em}
\subsubsection{Training and Latency Details.}
2K Retrofit is trained independently from the baseline models, while the loss function is inherited from the corresponding base model. Training proceeds for 13 epochs with a batch size of 8, requiring approximately a day on two NVIDIA RTX A6000 GPUs. Inference latency is evaluated on a single RTX 4090 using FP16 precision. 

\vspace{-1em}
\begin{figure}
\centering
\includegraphics[height=6.5cm,width=0.98\linewidth]{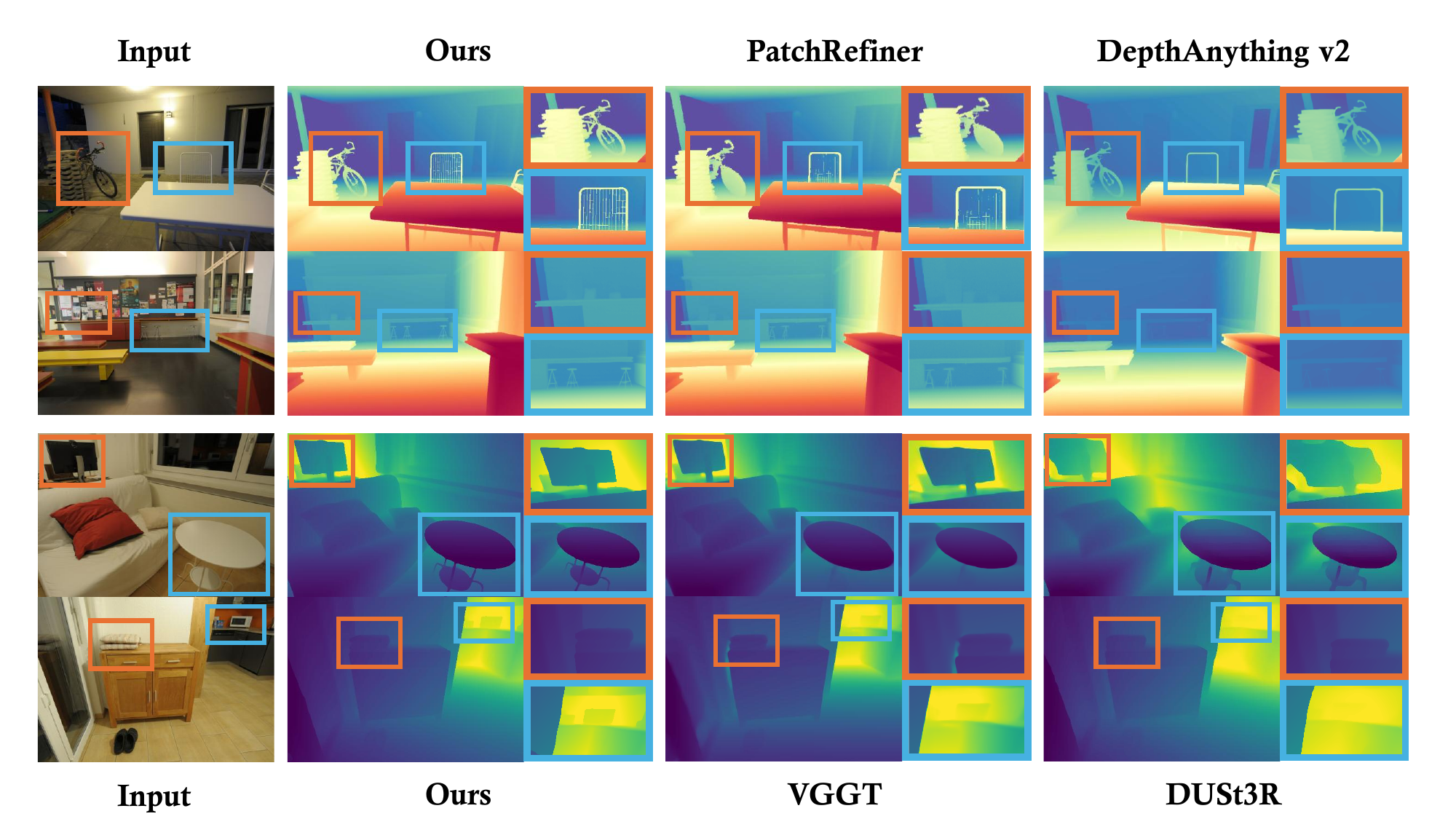}
\caption{Qualitative comparisons of 2K Retrofit with state-of-the-art methods on the ETH3D dataset. 
The top two rows show results on the \textbf{depth estimation} task, while the bottom two rows present results on the \textbf{point map estimation} task. 2K Retrofit consistently produces sharper high-resolution geometry and recovers fine structures more accurately than competing methods.
}\label{fig4}
\end{figure}

\vspace{-3em}

\subsection{Monocular 2K Depth Estimation}

We first evaluate 2K Retrofit on monocular 2K depth estimation to verify its ability of enhance high-resolution geometric prediction without retraining the base model. Monocular depth estimation serves as a fundamental benchmark for geometric perception, providing a direct and quantitative way to assess a model’s ability to recover fine-grained 3D details from high-resolution inputs. Depth Anything~v2~\cite{da2} is used as the frozen backbone. Given a single high-resolution RGB image, our method produces a dense 2K depth map by combining a coarse low-resolution prediction with entropy-guided sparse refinement.

Experiments are conducted on the 2K Retrofit synthetic dataset and two real-world benchmarks—ScanNet++~\cite{scannet++} and ARKitScenes~\cite{arkit}—both providing $1440\times1920$ depth annotations. Following standard protocols, we use their official training/validation splits, and additionally test on ETH3D~\cite{ETH} in a zero-shot setting to assess generalization. Performance is measured using standard monocular depth metrics: AbsRel, $\delta_{0.5}$, and RMSE.

\noindent\textbf{Results.} We compare 2K Retrofit with state-of-the-art (SOTA) depth estimation methods from two categories: monocular depth estimation (MDE) and depth completion/upsampling. For MDE, we include Depth Anything~v2~\cite{da2} and Marigold~\cite{marigold}; for completion and upsampling, we evaluate BPNet~\cite{bpnet}, Depth Prompting~\cite{DP}, MSPF~\cite{mspf}, BoostingDepth~\cite{boosting}, PatchFusion~\cite{patchfusion}, PatchRefiner~\cite{patchrefiner}, PRO~\cite{pro}, and PromptDA~\cite{promptda}. 
All methods are evaluated under two settings: \textit{zero-shot} (trained only on our synthetic 2K Retrofit dataset) and \textit{non zero-shot}.

As shown in Table~\ref{tab1}, Table~\ref{tab2}, and Fig.~\ref{fig4}, 2K Retrofit achieves the best overall performance on ARKitScenes, ScanNet++, and ETH3D. Compared with the strongest baseline PRO~\cite{pro}, our method reduces AbsRel by over 30\% while running more than three times faster (8.1 vs.\ 6.2 FPS), demonstrating advantages in both accuracy and efficiency.

Qualitative comparisons in Fig.~\ref{fig4} show that 2K Retrofit reconstructs sharper object boundaries and cleaner thin structures, especially near occlusions and high-frequency regions. These improvements stem from entropy-guided sparse refinement, which focuses computation on uncertain regions while avoiding redundant processing.

\subsection{Multi-View 2K Pointmap Estimation}
We further evaluate 2K Retrofit on high-resolution multi-view 3D pointmap estimation to examine its generalization beyond monocular depth prediction. Multi-view pointmap estimation serves as a more comprehensive benchmark for geometric reasoning, as it requires consistent 3D reconstruction across multiple views rather than single frame depth prediction. VGGT~\cite{vggt} is adopted as the frozen backbone. Given a set of multi-view RGB images, our framework refines the coarse pointmaps produced by the backbone through entropy-guided sparse correction, enabling accurate and consistent 2K-level 3D reconstruction.

\begin{table}[t]
    \centering
    \begin{subtable}{0.52\linewidth}
        \centering
        \scalebox{0.6}{
        \begin{tabular}{c|cccc}
        \toprule
        \multirow{2}{*}{\raisebox{-0.4ex}{Method}}& \multicolumn{4}{c}{Monocular Depth} \\
        \cmidrule{2-5}
        & AbsRel $\downarrow$ & RMSE $\downarrow$ & $\delta_{0.5}$ $\uparrow$ & FPS \\
        \midrule
        BoostingDepth~\cite{boosting} (CVPR 2021)& 0.0375 & 0.1201 & 0.9503 & 5.0\\
        DepthAnything v2*~\cite{da2} (NeurIPS 2024) & 0.0342 & 0.1129 & 0.9550 & 4.7 \\
        Marigold~\cite{marigold} (CVPR 2024)& 0.0711 & 0.1305 & 0.8849 & 1.3  \\
        BPNet~\cite{bpnet} (CVPR 2024)& 0.1494 & 0.2561 & 0.9137 & 2.6 \\
        PatchFusion$_{P=177}$~\cite{patchfusion} (CVPR 2024)& 0.0255 & 0.1104 & 0.9618 & 3.0\\
        PatchRefiner~\cite{patchrefiner} (ECCV 2024)& 0.0240 & 0.0987 & 0.9659 & 3.1\\
        PRO~\cite{pro} (ICCV 2025) & 0.0217 & 0.0902 & 0.9688 & 6.2\\
        \midrule
        2K Retrofit (Ours) & \textbf{0.0192} & \textbf{0.0877}  & \textbf{0.9700}  &  \textbf{8.1}  \\
        \bottomrule
        \end{tabular}}
        \caption{Monocular depth estimation.}
        \label{tab2a}
    \end{subtable}
    \hfill
    \begin{subtable}{0.45\linewidth}
        \centering
        \scalebox{0.6}{
        \begin{tabular}{c|cccc}
        \toprule
        \multirow{2}{*}{\raisebox{-0.4ex}{Method}}& \multicolumn{4}{c}{Point Map Estimation} \\
        \cmidrule{2-5}
        & Acc. $\downarrow$ &  Comp. $\downarrow$ &  Overall $\downarrow$ & FPS \\
        \midrule
        DUSt3R~\cite{dust3r} (CVPR 2024)& 1.462 &  0.927 & 1.195 & 1.1\\
        MASt3R~\cite{mast3r} (ECCV 2024)& 1.267 & 0.784 & 1.026  & 1.8\\
        VGGT~\cite{vggt} (CVPR 2025) & 1.167 & 0.639 & 0.903 & 2.6\\
        WinT3R~\cite{wint3r} (ArXiv 2025)& 1.131 & 0.617 & 0.891 & 3.2\\
        StreamVGGT~\cite{streamvggt} (ICLR 2026)& 1.140 & 0.613 & 0.870 & 3.5\\
        \midrule
        2K Retrofit (Ours) & \textbf{0.935} & \textbf{0.602}  & \textbf{0.839}  &  \textbf{5.5}  \\
        \bottomrule
        \end{tabular}}
        \caption{Point map estimation.}
        \label{tab2b}
    \end{subtable}

    \caption{Quantitative comparisons on ETH3D~\cite{ETH} dataset.}
    \label{tab2}
\vspace{-2em}
\end{table}

Experiments are conducted on both synthetic (MVS-Synth~\cite{mvssyn} and our 2K Retrofit synthetic dataset) and real-world datasets (DL3DV~\cite{dl3dv} and BlendMVS~\cite{blendedmvs}), with an additional zero-shot evaluation on ETH3D~\cite{ETH}. We follow standard pointmap metrics~\cite{pmetric}: average accuracy, average completeness, and the overall average error.

\noindent\textbf{Results.} We benchmark 2K Retrofit against state-of-the-art feedforward 3D reconstruction models, including DUSt3R~\cite{dust3r}, MASt3R~\cite{mast3r}, WinT3R~\cite{wint3r}, StreamVGGT~\cite{streamvggt}, and VGGT~\cite{vggt}. As shown in Table~\ref{tab2}, our method consistently improves pointmap estimation on ETH3D. Compared with the strongest baseline VGGT~\cite{vggt}, 2K Retrofit reduces average accuracy error and completeness by roughly 20\% and 6\%, respectively, yielding an overall error reduction of about 15\%. These results demonstrate that selective refinement significantly improves geometric precision without retraining the backbone.

Qualitative comparisons in Fig.~\ref{fig4} further support these findings. 2K Retrofit reconstructs finer structures and cleaner surfaces, particularly around thin objects and occlusion boundaries where foundation models typically degrade at 2K resolution. 

In addition to accuracy gains, 2K Retrofit runs at 5.5~FPS—over 3$\times$ faster than VGGT (2.6~FPS) and nearly 8$\times$ faster than DUSt3R (1.1~FPS)—showing a favorable balance between accuracy and computational efficiency for high-resolution 3D reconstruction.

We further compare our method with multi-view stereo (MVS) approaches. We note that many MVS methods, such as RAFT-Stereo~\cite{raft} and MVSFormer++~\cite{mvsformer++}, also follow a coarse-to-fine paradigm. Therefore, we conduct additional evaluations on the ETH3D benchmark. For fairness, MVSFormer++~\cite{mvsformer++} is evaluated with both its original weights and weights fine-tuned on our synthetic dataset, denoted as MVSFormer++$^{\dagger}$. Results in Table~\ref{tabmvs} show that our method outperforms these approaches and remains competitive with MVSFormer++$^{\dagger}$, while being substantially faster.

\begin{table}[h]
\vspace{-1em}
\centering
\scalebox{0.8}{
\begin{tabular}{lccc}
\toprule
Methods & Precision$\uparrow$ & F1-Score$\uparrow$ & Runtime(s)$\downarrow$\\
\midrule
CascadeMVS~\cite{cascade} (CVPR2020)    & 79.17 & 80.84 & 0.4 \\
RAFT-Stereo~\cite{raft} (3DV 2021)      & 80.36 & 81.42 & 0.8 \\
MVSFormer++~\cite{mvsformer++} (ICLR2024)     & 82.23 & 83.75 & 1.5 \\
MVSFormer++$^{\dagger}$~\cite{mvsformer++} (ICLR2024)       & \textbf{84.49} & \textbf{85.13} & 1.5 \\
\midrule
2K Retrofit (Ours)                  & 84.01 & 84.88 & \textbf{0.3} \\
\bottomrule
\end{tabular}
}
\caption{Dense MVS Estimation on the ETH~\cite{ETH} Dataset.}
\label{tabmvs}
\vspace{-4em}
\end{table}

\subsection{Efficiency Analysis on ETH3D.}
We compare the efficiency of 2K Retrofit with retraining VGGT at 2K resolution (Table~\ref{tab:eth3d_efficiency_final}). Full retraining requires 76.5\,GB memory and 495\,GFLOPs per forward pass, whereas our method reduces these costs by over 50\% (32.8\,GB, 172\,GFLOPs). While retrained VGGT achieves slightly higher accuracy, 2K Retrofit runs at 8.5\,FPS versus 0.5\,FPS, providing a 17$\times$ speedup with competitive performance, demonstrating a practical balance between accuracy and efficiency for high-resolution 3D geometry prediction.

\begin{table}[h]
\vspace{-1em}
\centering
\resizebox{\columnwidth}{!}{
\begin{tabular}{lcccccc}
\toprule
\textbf{Method} & \textbf{VRAM (GB) $\downarrow$} & \textbf{GFLOPs $\downarrow$} & \textbf{Accuracy $\downarrow$} & \textbf{Completeness $\downarrow$} & \textbf{FPS $\uparrow$} \\
\midrule
VGGT retrained at 2K & 76.5 & 495 & \textbf{0.911} & \textbf{0.593} & 0.8 \\
2K Retrofit (Ours) & \textbf{32.8} & \textbf{172} & 0.935 & 0.602 & \textbf{5.5} \\
\bottomrule
\end{tabular}
}
\caption{\textbf{Efficiency comparison on ETH3D point map estimation task.}}
\label{tab:eth3d_efficiency_final}
\vspace{-4em}
\end{table}

\subsection{Generalization and Robustness Analysis}
\begin{figure}[h]
\centering

\begin{subfigure}{\linewidth}
    \centering
    \includegraphics[width=0.98\linewidth,height=2cm]{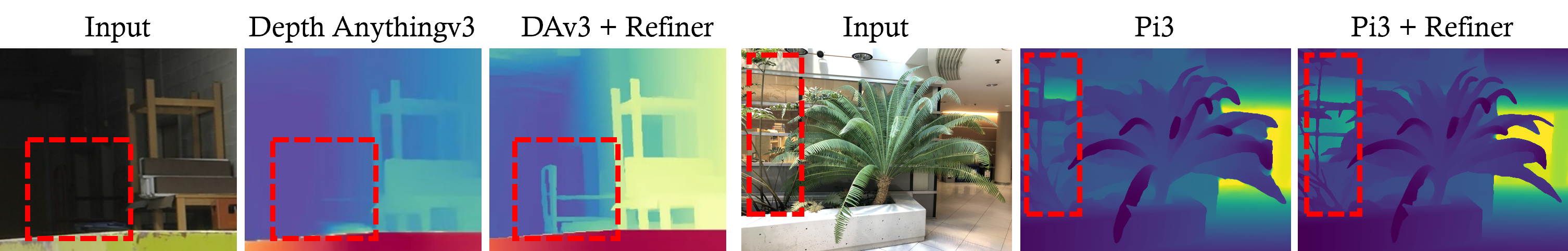}
    \caption{Generality across different backbones.}
    \label{fig:oool_b}
\end{subfigure}

\begin{subfigure}{\linewidth}
    \centering
    \includegraphics[width=0.98\linewidth,height=2.5cm]{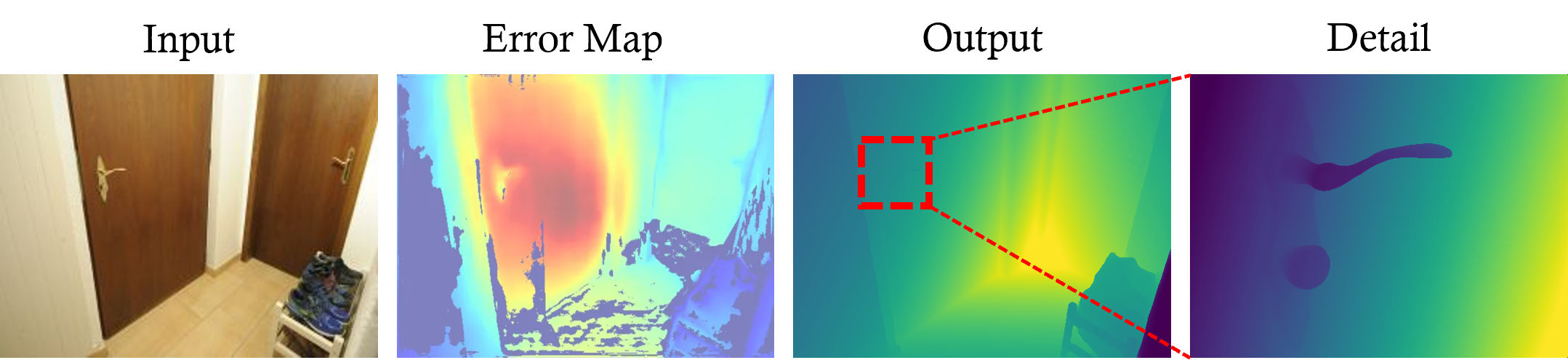}
    \caption{Robustness visualization under challenging condition.}
    \label{fig:oool_a}
\end{subfigure}

\caption{Generalization and Robustness Analysis.}
\label{fig:oool}

\end{figure}

To further evaluate the generalization and robustness of our approach, we extend our experiments to additional monocular depth models, including Pi3~\cite{pi3} and Depth Anything v3~\cite{dav3} (Fig.~\ref{fig:oool_b}). These models are tested without retraining, demonstrating the plug-and-play compatibility of our framework across different architectures and consistently improving reconstruction quality. 

We also present representative failure cases in Fig.~\ref{fig:oool_a}. Our method may still struggle in extremely textureless or highly reflective regions—common challenges in depth estimation and 3D reconstruction—where errors can become dense. Nevertheless, our approach remains capable of refining high-frequency structural details in most regions, resulting in improved geometric fidelity.

\vspace{-1em}

\vspace{-2.5em}

\subsection{Ablation Study}
We analyze alternative designs for key components of our method, focusing on point map estimation. Detailed breakdowns of accuracy and efficiency improvements are provided below. 

\noindent\textbf{Impact of Entropy Selectors.} 
We compare our entropy-based pixel selector with several alternatives, including a random selector, a learnable selector, edge-based heuristics, and bilateral upsampling (Table~\ref{tab:pixel_selector}). 
The random selector fails to surpass the low-resolution baseline, showing that unguided refinement is ineffective at 2K resolution. 
The learnable selector slightly improves recall but nearly doubles latency and memory. 
Edge-based heuristics and bilateral upsampling perform better but still lag behind entropy-guided selection. Overall, entropy-based selection achieves the best accuracy–efficiency trade-off.


\begin{table}[h]
\centering
\begin{subtable}[t]{0.48\linewidth}
\centering
\scalebox{0.8}{
\begin{tabular}{l|ccc}
\toprule
Pixel Selector & Acc.\,$\downarrow$ & Comp.\,$\downarrow$ & FPS \\
\midrule
Random   & 1.126 & 0.835 & 4.1\\
Edge-based heuristics   & 0.942 & 0.625 & 4.1\\
Bilateral upsampling & 0.940 & 0.617 & 4.0\\
Learnable & \textbf{0.921} & 0.635 & 3.8\\
\midrule
Entropy & 0.935 & \textbf{0.602} & \textbf{5.5}\\
\bottomrule
\end{tabular}}
\subcaption{The pixel selector.}
\label{tab:pixel_selector}
\end{subtable}
\begin{subtable}[t]{0.48\linewidth}
\centering
\begin{tabular}{l|ccc}
\toprule
Fusion strategy & Acc.\,$\downarrow$ & Comp.\,$\downarrow$ & FPS \\
\midrule
Direct  & 1.113 & 0.735 & \textbf{5.9}\\
Entropy & 0.940 & 0.607 & 5.0\\
\midrule
Gated & \textbf{0.935} & \textbf{0.602} & 5.5\\
\bottomrule
\end{tabular}
\subcaption{The Fusion strategy.}
\label{tab:fusion}
\end{subtable}%

\begin{subtable}[t]{0.48\linewidth}
\centering
\scalebox{0.8}{
\begin{tabular}{l|ccc}
\toprule
$\alpha$ & Acc.\,$\downarrow$ & Comp.\,$\downarrow$ & FPS\\
\midrule
0.8 & 1.135 & 0.633 & 8.0\\
0.6 & 0.962 & 0.621 & 7.3\\
0.1 & \textbf{0.917} & \textbf{0.595} & 3.4\\
\midrule
0.3 & 0.935 & 0.602 & \textbf{5.5}\\
\bottomrule
\end{tabular}}
\subcaption{The entropy threshold.}
\label{tab:entropy_thry}
\end{subtable}%
\hfill
\begin{subtable}[t]{0.48\linewidth}
\centering
\scalebox{0.7}{
\begin{tabular}{l|ccc}
\toprule
Sparse Feature Extractor & Acc.\,$\downarrow$ & Comp.\,$\downarrow$ & FPS\\
\midrule
Point Transformer   & \textbf{0.919} & \textbf{0.587} & 1.7\\
\midrule
MinkowskiUNet   & 0.935 & 0.602 & \textbf{5.5} \\
\bottomrule
\end{tabular}}
\subcaption{The sparse feature extractor.}
\label{tab:sparse}
\end{subtable}%
\caption{Ablation experiments to validate our design choices.}
\label{tab:selector_all}
\vspace{-1cm}
\end{table}

\noindent\textbf{Impact of Fusion Strategies.}
As shown in Table~\ref{tab:fusion}, direct replacement performs poorly due to the loss of global context, while entropy-based fusion improves stability but still lacks feature-level adaptivity. 
Our gated fusion achieves the lowest errors (Acc.\,0.935, Comp.\,0.602) with comparable FPS, confirming its effectiveness in selectively integrating coarse and refined estimates.

\noindent\textbf{Impact of Entropy Thresholds.} 
We further study the influence of the entropy threshold $\alpha$ on accuracy and latency (see Table~\ref{tab:entropy_thry}). Lowering the threshold increases the number of selected pixels, leading to better reconstruction accuracy but also higher latency. In contrast, higher thresholds reduce computational load at the expense of finer details. We choose $\alpha = 0.3$ as a moderate setting that achieves strong accuracy while maintaining the largest speedup.

\noindent\textbf{Impact of Sparse Extractor Capacity.} 
We also evaluate different backbone choices for sparse feature extraction (Table~\ref{tab:sparse}). Using a point transformer as the sparse feature extractor yields marginally higher accuracy but considerably reduces inference speed. In contrast, MinkowskiUNet offers an optimal balance of efficiency and accuracy, and is therefore adopted as our default extractor.

\section{Conclusion} 
\label{sec:conclusion} 
This paper proposes 2K Retrofit, a general framework designed to overcome the scalability limitations of current geometric foundation models for high-resolution prediction. By combining fast coarse inference with entropy-based sparse refinement, our method efficiently bridges the gap between low-resolution backbones and 2K-level outputs. Unlike previous approaches that require network modification, patch-wise processing, or retraining, 2K Retrofit directly adapts existing foundation models to high-resolution scenarios without altering their original architectures. This modular and easily integrable design makes the method broadly applicable to a range of geometric prediction tasks, including depth estimation and 3D reconstruction. Extensive experiments on both 2K depth estimation and 2K pointmap reconstruction demonstrate that 2K Retrofit achieves superior accuracy and significantly higher efficiency compared with existing methods.

\paragraph{Limitations.}
2K Retrofit may still struggle in challenging scenarios such as reflective surfaces, transparent objects, and low-texture regions, where geometric cues are inherently ambiguous, leaving a future research direction of incorporating stronger geometric priors into our model.

\bibliographystyle{splncs04}

\begin{thebibliography}{10}
\providecommand{\url}[1]{\texttt{#1}}
\providecommand{\urlprefix}{URL }
\providecommand{\doi}[1]{https://doi.org/#1}

\bibitem{55}
Abdar, M., Pourpanah, F., Hussain, S., Rezazadegan, D., Liu, L., Ghavamzadeh, M., Fieguth, P., Cao, X., Khosravi, A., Acharya, U.R., et~al.: A review of uncertainty quantification in deep learning: Techniques, applications and challenges. Information fusion  \textbf{76},  243--297 (2021)

\bibitem{zoedepth}
Bhat, S.F., Birkl, R., Wofk, D., Wonka, P., M{\"u}ller, M.: Zoedepth: Zero-shot transfer by combining relative and metric depth. arXiv preprint arXiv:2302.12288  (2023)

\bibitem{15}
Bhat, S.F., Birkl, R., Wofk, D., Wonka, P., M{\"u}ller, M.: Zoedepth: Zero-shot transfer by combining relative and metric depth. arXiv preprint arXiv:2302.12288  (2023)

\bibitem{rel}
Bian, J.W., Zhan, H., Wang, N., Chin, T.J., Shen, C., Reid, I.: Auto-rectify network for unsupervised indoor depth estimation. IEEE transactions on pattern analysis and machine intelligence  \textbf{44}(12),  9802--9813 (2021)

\bibitem{depthpro}
Bochkovskii, A., Delaunoy, A., Germain, H., Santos, M., Zhou, Y., Richter, S.R., Koltun, V.: Depth pro: Sharp monocular metric depth in less than a second. arXiv preprint arXiv:2410.02073  (2024)

\bibitem{42}
Bolya, D., Fu, C.Y., Dai, X., Zhang, P., Feichtenhofer, C., Hoffman, J.: Token merging: Your vit but faster. arXiv preprint arXiv:2210.09461  (2022)

\bibitem{43}
Bolya, D., Hoffman, J.: Token merging for fast stable diffusion. In: Proceedings of the IEEE/CVF conference on computer vision and pattern recognition. pp. 4599--4603 (2023)

\bibitem{mvsformer}
Cao, C., Ren, X., Fu, Y.: Mvsformer: Multi-view stereo by learning robust image features and temperature-based depth. arXiv preprint arXiv:2208.02541  (2022)

\bibitem{mvsformer++}
Cao, C., Ren, X., Fu, Y.: Mvsformer++: Revealing the devil in transformer's details for multi-view stereo. arXiv preprint arXiv:2401.11673  (2024)

\bibitem{45}
Choy, C., Gwak, J., Savarese, S.: 4d spatio-temporal convnets: Minkowski convolutional neural networks. In: Proceedings of the IEEE/CVF conference on computer vision and pattern recognition. pp. 3075--3084 (2019)

\bibitem{9}
Eigen, D., Fergus, R.: Predicting depth, surface normals and semantic labels with a common multi-scale convolutional architecture. In: Proceedings of the IEEE international conference on computer vision. pp. 2650--2658 (2015)

\bibitem{2}
Eigen, D., Puhrsch, C., Fergus, R.: Depth map prediction from a single image using a multi-scale deep network. Advances in neural information processing systems  \textbf{27} (2014)

\bibitem{10}
Eigen, D., Puhrsch, C., Fergus, R.: Depth map prediction from a single image using a multi-scale deep network. Advances in neural information processing systems  \textbf{27} (2014)

\bibitem{dense3r}
Fang, X., Gao, J., Wang, Z., Chen, Z., Ren, X., Lyu, J., Ren, Q., Yang, Z., Yang, X., Yan, Y., et~al.: Dens3r: A foundation model for 3d geometry prediction. arXiv preprint arXiv:2507.16290  (2025)

\bibitem{26}
Fu, Q., Xu, Q., Ong, Y.S., Tao, W.: Geo-neus: Geometry-consistent neural implicit surfaces learning for multi-view reconstruction. Advances in Neural Information Processing Systems  \textbf{35},  3403--3416 (2022)

\bibitem{21}
Furukawa, Y., Hern{\'a}ndez, C., et~al.: Multi-view stereo: A tutorial. Foundations and trends{\textregistered} in Computer Graphics and Vision  \textbf{9}(1-2),  1--148 (2015)

\bibitem{22}
Galliani, S., Lasinger, K., Schindler, K.: Massively parallel multiview stereopsis by surface normal diffusion. In: Proceedings of the IEEE international conference on computer vision. pp. 873--881 (2015)

\bibitem{35}
Gao, P., Ma, T., Li, H., Lin, Z., Dai, J., Qiao, Y.: Convmae: Masked convolution meets masked autoencoders. arXiv preprint arXiv:2205.03892  (2022)

\bibitem{49}
Gondimalla, A., Chesnut, N., Thottethodi, M., Vijaykumar, T.: Sparten: A sparse tensor accelerator for convolutional neural networks. In: Proceedings of the 52nd Annual IEEE/ACM International Symposium on Microarchitecture. pp. 151--165 (2019)

\bibitem{27}
Gu, X., Fan, Z., Zhu, S., Dai, Z., Tan, F., Tan, P.: Cascade cost volume for high-resolution multi-view stereo and stereo matching. In: Proceedings of the IEEE/CVF conference on computer vision and pattern recognition. pp. 2495--2504 (2020)

\bibitem{cascade}
Gu, X., Fan, Z., Zhu, S., Dai, Z., Tan, F., Tan, P.: Cascade cost volume for high-resolution multi-view stereo and stereo matching. In: Proceedings of the IEEE/CVF conference on computer vision and pattern recognition. pp. 2495--2504 (2020)

\bibitem{33}
Guo, M.H., Lu, C.Z., Hou, Q., Liu, Z., Cheng, M.M., Hu, S.M.: Segnext: Rethinking convolutional attention design for semantic segmentation. arxiv. 2022. arXiv preprint arXiv:2209.08575  (2022)

\bibitem{6}
Han, S.H., Park, M.G., Yoon, J.H., Kang, J.M., Park, Y.J., Jeon, H.G.: High-fidelity 3d human digitization from single 2k resolution images. In: Proceedings of the IEEE/CVF Conference on Computer Vision and Pattern Recognition. pp. 12869--12879 (2023)

\bibitem{36}
He, K., Chen, X., Xie, S., Li, Y., Doll{\'a}r, P., Girshick, R.: Masked autoencoders are scalable vision learners. In: Proceedings of the IEEE/CVF conference on computer vision and pattern recognition. pp. 16000--16009 (2022)

\bibitem{46}
Hong, K., Yu, Z., Dai, G., Yang, X., Lian, Y., Xu, N., Wang, Y.: Exploiting hardware utilization and adaptive dataflow for efficient sparse convolution in 3d point clouds. Proceedings of Machine Learning and Systems  \textbf{5},  428--441 (2023)

\bibitem{37}
Huang, L., You, S., Zheng, M., Wang, F., Qian, C., Yamasaki, T.: Green hierarchical vision transformer for masked image modeling. Advances in Neural Information Processing Systems  \textbf{35},  19997--20010 (2022)

\bibitem{mvssyn}
Huang, P.H., Matzen, K., Kopf, J., Ahuja, N., Huang, J.B.: Deepmvs: Learning multi-view stereopsis. In: Proceedings of the IEEE conference on computer vision and pattern recognition. pp. 2821--2830 (2018)

\bibitem{56}
Huang, Y.H., Proesmans, M., Georgoulis, S., Van~Gool, L.: Uncertainty based model selection for fast semantic segmentation. In: 2019 16th International Conference on Machine Vision Applications (MVA). pp.~1--6. IEEE (2019)

\bibitem{arkit}
Huang, Y.H., Proesmans, M., Georgoulis, S., Van~Gool, L.: Uncertainty based model selection for fast semantic segmentation. In: 2019 16th International Conference on Machine Vision Applications (MVA). pp.~1--6. IEEE (2019)

\bibitem{16}
Hui, T.W., Loy, C.C., Tang, X.: Depth map super-resolution by deep multi-scale guidance. In: European conference on computer vision. pp. 353--369. Springer (2016)

\bibitem{mvsanywhere}
Izquierdo, S., Sayed, M., Firman, M., Garcia-Hernando, G., Turmukhambetov, D., Civera, J., Mac~Aodha, O., Brostow, G., Watson, J.: Mvsanywhere: Zero-shot multi-view stereo. In: Proceedings of the Computer Vision and Pattern Recognition Conference. pp. 11493--11504 (2025)

\bibitem{marigold}
Ke, B., Obukhov, A., Huang, S., Metzger, N., Daudt, R.C., Schindler, K.: Repurposing diffusion-based image generators for monocular depth estimation. In: Proceedings of the IEEE/CVF conference on computer vision and pattern recognition. pp. 9492--9502 (2024)

\bibitem{52}
Kirillov, A., Wu, Y., He, K., Girshick, R.: Pointrend: Image segmentation as rendering. In: Proceedings of the IEEE/CVF conference on computer vision and pattern recognition. pp. 9799--9808 (2020)

\bibitem{38}
Kong, Z., Dong, P., Ma, X., Meng, X., Sun, M., Niu, W., Shen, X., Yuan, G., Ren, B., Qin, M., et~al.: Spvit: enabling faster vision transformers via soft token pruning (2022). URL https://arxiv. org/abs/2112.13890  (2021)

\bibitem{pro}
Kwon, B., Kim, M.: One look is enough: Seamless patchwise refinement for zero-shot monocular depth estimation on high-resolution images. In: Proceedings of the IEEE/CVF International Conference on Computer Vision. pp. 8077--8087 (2025)

\bibitem{mast3r}
Leroy, V., Cabon, Y., Revaud, J.: Grounding image matching in 3d with mast3r. In: European Conference on Computer Vision. pp. 71--91. Springer (2024)

\bibitem{11}
Li, Z., Snavely, N.: Megadepth: Learning single-view depth prediction from internet photos. In: Proceedings of the IEEE conference on computer vision and pattern recognition. pp. 2041--2050 (2018)

\bibitem{patchfusion}
Li, Z., Bhat, S.F., Wonka, P.: Patchfusion: An end-to-end tile-based framework for high-resolution monocular metric depth estimation. In: Proceedings of the IEEE/CVF Conference on Computer Vision and Pattern Recognition. pp. 10016--10025 (2024)

\bibitem{patchrefiner}
Li, Z., Bhat, S.F., Wonka, P.: Patchrefiner: Leveraging synthetic data for real-domain high-resolution monocular metric depth estimation. In: European Conference on Computer Vision. pp. 250--267. Springer (2024)

\bibitem{wint3r}
Li, Z., Zhou, J., Wang, Y., Guo, H., Chang, W., Zhou, Y., Zhu, H., Chen, J., Shen, C., He, T.: Wint3r: Window-based streaming reconstruction with camera token pool. arXiv preprint arXiv:2509.05296  (2025)

\bibitem{39}
Liang, Y., Ge, C., Tong, Z., Song, Y., Wang, J., Xie, P.: Not all patches are what you need: Expediting vision transformers via token reorganizations. arXiv preprint arXiv:2202.07800  (2022)

\bibitem{dav3}
Lin, H., Chen, S., Liew, J., Chen, D.Y., Li, Z., Shi, G., Feng, J., Kang, B.: Depth anything 3: Recovering the visual space from any views. arXiv preprint arXiv:2511.10647  (2025)

\bibitem{promptda}
Lin, H., Peng, S., Chen, J., Peng, S., Sun, J., Liu, M., Bao, H., Feng, J., Zhou, X., Kang, B.: Prompting depth anything for 4k resolution accurate metric depth estimation. In: Proceedings of the Computer Vision and Pattern Recognition Conference. pp. 17070--17080 (2025)

\bibitem{50}
Lin, Y., Zhang, Z., Tang, H., Wang, H., Han, S.: Pointacc: Efficient point cloud accelerator. In: MICRO-54: 54th Annual IEEE/ACM International Symposium on Microarchitecture. pp. 449--461 (2021)

\bibitem{dl3dv}
Ling, L., Sheng, Y., Tu, Z., Zhao, W., Xin, C., Wan, K., Yu, L., Guo, Q., Yu, Z., Lu, Y., et~al.: Dl3dv-10k: A large-scale scene dataset for deep learning-based 3d vision. In: Proceedings of the IEEE/CVF Conference on Computer Vision and Pattern Recognition. pp. 22160--22169 (2024)

\bibitem{raft}
Lipson, L., Teed, Z., Deng, J.: Raft-stereo: Multilevel recurrent field transforms for stereo matching. In: 2021 International conference on 3D vision (3DV). pp. 218--227. IEEE (2021)

\bibitem{34}
Liu, J., Chen, Y., Ye, X., Tian, Z., Tan, X., Qi, X.: Spatial pruned sparse convolution for efficient 3d object detection. Advances in neural information processing systems  \textbf{35},  6735--6748 (2022)

\bibitem{sparserefine}
Liu, Z., Zhang, Z., Khaki, S., Yang, S., Tang, H., Xu, C., Keutzer, K., Han, S.: Sparse refinement for efficient high-resolution semantic segmentation. In: European Conference on Computer Vision. pp. 108--127. Springer (2024)

\bibitem{5}
Lorensen, W.E., Cline, H.E.: Marching cubes: A high resolution 3d surface construction algorithm. In: Seminal graphics: pioneering efforts that shaped the field, pp. 347--353 (1998)

\bibitem{sgdr}
Loshchilov, I., Hutter, F.: Sgdr: Stochastic gradient descent with warm restarts. arXiv preprint arXiv:1608.03983  (2016)

\bibitem{adamw}
Loshchilov, I., Hutter, F.: Decoupled weight decay regularization. arXiv preprint arXiv:1711.05101  (2017)

\bibitem{44}
Ma, X., Zhou, Y., Wang, H., Qin, C., Sun, B., Liu, C., Fu, Y.: Image as set of points. arXiv preprint arXiv:2303.01494  (2023)

\bibitem{28}
Ma, Z., Teed, Z., Deng, J.: Multiview stereo with cascaded epipolar raft. In: European Conference on Computer Vision. pp. 734--750. Springer (2022)

\bibitem{17}
Metzger, N., Daudt, R.C., Schindler, K.: Guided depth super-resolution by deep anisotropic diffusion. In: Proceedings of the IEEE/CVF Conference on Computer Vision and Pattern Recognition. pp. 18237--18246 (2023)

\bibitem{19}
Miangoleh, S.M.H., Dille, S., Mai, L., Paris, S., Aksoy, Y.: Boosting monocular depth estimation models to high-resolution via content-adaptive multi-resolution merging. In: Proceedings of the IEEE/CVF Conference on Computer Vision and Pattern Recognition. pp. 9685--9694 (2021)

\bibitem{boosting}
Miangoleh, S.M.H., Dille, S., Mai, L., Paris, S., Aksoy, Y.: Boosting monocular depth estimation models to high-resolution via content-adaptive multi-resolution merging. In: Proceedings of the IEEE/CVF Conference on Computer Vision and Pattern Recognition. pp. 9685--9694 (2021)

\bibitem{25}
Niemeyer, M., Mescheder, L., Oechsle, M., Geiger, A.: Differentiable volumetric rendering: Learning implicit 3d representations without 3d supervision. In: Proceedings of the IEEE/CVF conference on computer vision and pattern recognition. pp. 3504--3515 (2020)

\bibitem{3}
Ouyang, W., Song, X., Feng, B., Xu, Z.: Octocc: High-resolution 3d occupancy prediction with octree. In: Proceedings of the AAAI Conference on Artificial Intelligence. vol.~38, pp. 4369--4377 (2024)

\bibitem{31}
Pan, B., Lin, W., Fang, X., Huang, C., Zhou, B., Lu, C.: Recurrent residual module for fast inference in videos. In: Proceedings of the IEEE Conference on Computer Vision and Pattern Recognition. pp. 1536--1545 (2018)

\bibitem{32}
Pan, B., Panda, R., Fosco, C., Lin, C.C., Andonian, A., Meng, Y., Saenko, K., Oliva, A., Feris, R.: Va-red $^2$: Video adaptive redundancy reduction. arXiv preprint arXiv:2102.07887  (2021)

\bibitem{40}
Pan, B., Panda, R., Jiang, Y., Wang, Z., Feris, R., Oliva, A.: Ia-red $^2$: Interpretability-aware redundancy reduction for vision transformers. Advances in neural information processing systems  \textbf{34},  24898--24911 (2021)

\bibitem{DP}
Park, J.H., Jeong, C., Lee, J., Jeon, H.G.: Depth prompting for sensor-agnostic depth estimation. In: Proceedings of the IEEE/CVF Conference on Computer Vision and Pattern Recognition. pp. 9859--9869 (2024)

\bibitem{29}
Peng, R., Wang, R., Wang, Z., Lai, Y., Wang, R.: Rethinking depth estimation for multi-view stereo: A unified representation. In: Proceedings of the IEEE/CVF conference on computer vision and pattern recognition. pp. 8645--8654 (2022)

\bibitem{54}
Qi, C.R., Su, H., Mo, K., Guibas, L.J.: Pointnet: Deep learning on point sets for 3d classification and segmentation. In: Proceedings of the IEEE conference on computer vision and pattern recognition. pp. 652--660 (2017)

\bibitem{dpt}
Ranftl, R., Bochkovskiy, A., Koltun, V.: Vision transformers for dense prediction. In: Proceedings of the IEEE/CVF international conference on computer vision. pp. 12179--12188 (2021)

\bibitem{13}
Ranftl, R., Lasinger, K., Hafner, D., Schindler, K., Koltun, V.: Towards robust monocular depth estimation: Mixing datasets for zero-shot cross-dataset transfer. IEEE transactions on pattern analysis and machine intelligence  \textbf{44}(3),  1623--1637 (2020)

\bibitem{41}
Rao, Y., Zhao, W., Liu, B., Lu, J., Zhou, J., Hsieh, C.J.: Dynamicvit: Efficient vision transformers with dynamic token sparsification. Advances in neural information processing systems  \textbf{34},  13937--13949 (2021)

\bibitem{47}
Ren, M., Pokrovsky, A., Yang, B., Urtasun, R.: Sbnet: Sparse blocks network for fast inference. In: Proceedings of the IEEE conference on Computer Vision and Pattern Recognition. pp. 8711--8720 (2018)

\bibitem{20}
Rey-Area, M., Yuan, M., Richardt, C.: 360monodepth: High-resolution 360deg monocular depth estimation. In: Proceedings of the IEEE/CVF Conference on Computer Vision and Pattern Recognition. pp. 3762--3772 (2022)

\bibitem{8}
Riegler, G., R{\"u}ther, M., Bischof, H.: Atgv-net: Accurate depth super-resolution. In: European conference on computer vision. pp. 268--284. Springer (2016)

\bibitem{14}
Rombach, R., Blattmann, A., Lorenz, D., Esser, P., Ommer, B.: High-resolution image synthesis with latent diffusion models. In: Proceedings of the IEEE/CVF conference on computer vision and pattern recognition. pp. 10684--10695 (2022)

\bibitem{23}
Sch{\"o}nberger, J.L., Zheng, E., Frahm, J.M., Pollefeys, M.: Pixelwise view selection for unstructured multi-view stereo. In: European conference on computer vision. pp. 501--518. Springer (2016)

\bibitem{ETH}
Schops, T., Schonberger, J.L., Galliani, S., Sattler, T., Schindler, K., Pollefeys, M., Geiger, A.: A multi-view stereo benchmark with high-resolution images and multi-camera videos. In: Proceedings of the IEEE conference on computer vision and pattern recognition. pp. 3260--3269 (2017)

\bibitem{57}
Spencer, J., Russell, C., Hadfield, S., Bowden, R.: Kick back \& relax: Learning to reconstruct the world by watching slowtv. In: Proceedings of the IEEE/CVF International Conference on Computer Vision. pp. 15768--15779 (2023)

\bibitem{7}
Sun, K., Zhao, Y., Jiang, B., Cheng, T., Xiao, B., Liu, D., Mu, Y., Wang, X., Liu, W., Wang, J.: High-resolution representations for labeling pixels and regions. arXiv preprint arXiv:1904.04514  (2019)

\bibitem{48}
Tang, H., Liu, Z., Li, X., Lin, Y., Han, S.: Torchsparse: Efficient point cloud inference engine. Proceedings of Machine Learning and Systems  \textbf{4},  302--315 (2022)

\bibitem{bpnet}
Tang, J., Tian, F.P., An, B., Li, J., Tan, P.: Bilateral propagation network for depth completion. In: Proceedings of the IEEE/CVF Conference on Computer Vision and Pattern Recognition. pp. 9763--9772 (2024)

\bibitem{4}
Tang, S., Chen, J., Wang, D., Tang, C., Zhang, F., Fan, Y., Chandra, V., Furukawa, Y., Ranjan, R.: Mvdiffusion++: A dense high-resolution multi-view diffusion model for single or sparse-view 3d object reconstruction. In: European Conference on Computer Vision. pp. 175--191. Springer (2024)

\bibitem{domain}
Tobin, J., Fong, R., Ray, A., Schneider, J., Zaremba, W., Abbeel, P.: Domain randomization for transferring deep neural networks from simulation to the real world. In: 2017 IEEE/RSJ international conference on intelligent robots and systems (IROS). pp. 23--30. IEEE (2017)

\bibitem{pmetric}
Umeyama, S.: Least-squares estimation of transformation parameters between two point patterns. IEEE Transactions on pattern analysis and machine intelligence  \textbf{13}(4),  376--380 (2002)

\bibitem{vggt}
Wang, J., Chen, M., Karaev, N., Vedaldi, A., Rupprecht, C., Novotny, D.: Vggt: Visual geometry grounded transformer. In: Proceedings of the Computer Vision and Pattern Recognition Conference. pp. 5294--5306 (2025)

\bibitem{12}
Wang, Q., Zheng, S., Yan, Q., Deng, F., Zhao, K., Chu, X.: Irs: A large naturalistic indoor robotics stereo dataset to train deep models for disparity and surface normal estimation. arXiv preprint arXiv:1912.09678  (2019)

\bibitem{dust3r}
Wang, S., Leroy, V., Cabon, Y., Chidlovskii, B., Revaud, J.: Dust3r: Geometric 3d vision made easy. In: Proceedings of the IEEE/CVF Conference on Computer Vision and Pattern Recognition. pp. 20697--20709 (2024)

\bibitem{pi3}
Wang, Y., Zhou, J., Zhu, H., Chang, W., Zhou, Y., Li, Z., Chen, J., Pang, J., Shen, C., He, T.: {$\pi^3$}: Permutation-equivariant visual geometry learning. arXiv preprint arXiv:2507.13347  (2025)

\bibitem{24}
Wei, Y., Liu, S., Rao, Y., Zhao, W., Lu, J., Zhou, J.: Nerfingmvs: Guided optimization of neural radiance fields for indoor multi-view stereo. In: Proceedings of the IEEE/CVF international conference on computer vision. pp. 5610--5619 (2021)

\bibitem{foundationstereo}
Wen, B., Trepte, M., Aribido, J., Kautz, J., Gallo, O., Birchfield, S.: Foundationstereo: Zero-shot stereo matching. In: Proceedings of the Computer Vision and Pattern Recognition Conference. pp. 5249--5260 (2025)

\bibitem{mspf}
Xian, C., Qian, K., Zhang, Z., Wang, C.C.: Multi-scale progressive fusion learning for depth map super-resolution. arXiv preprint arXiv:2011.11865  (2020)

\bibitem{da}
Yang, L., Kang, B., Huang, Z., Xu, X., Feng, J., Zhao, H.: Depth anything: Unleashing the power of large-scale unlabeled data. In: Proceedings of the IEEE/CVF conference on computer vision and pattern recognition. pp. 10371--10381 (2024)

\bibitem{da2}
Yang, L., Kang, B., Huang, Z., Zhao, Z., Xu, X., Feng, J., Zhao, H.: Depth anything v2. Advances in Neural Information Processing Systems  \textbf{37},  21875--21911 (2024)

\bibitem{30}
Yao, Y., Luo, Z., Li, S., Fang, T., Quan, L.: Mvsnet: Depth inference for unstructured multi-view stereo. In: Proceedings of the European conference on computer vision (ECCV). pp. 767--783 (2018)

\bibitem{blendedmvs}
Yao, Y., Luo, Z., Li, S., Zhang, J., Ren, Y., Zhou, L., Fang, T., Quan, L.: Blendedmvs: A large-scale dataset for generalized multi-view stereo networks. In: Proceedings of the IEEE/CVF conference on computer vision and pattern recognition. pp. 1790--1799 (2020)

\bibitem{scannet++}
Yeshwanth, C., Liu, Y.C., Nie{\ss}ner, M., Dai, A.: Scannet++: A high-fidelity dataset of 3d indoor scenes. In: Proceedings of the IEEE/CVF International Conference on Computer Vision. pp. 12--22 (2023)

\bibitem{53}
Zhang, G., Lu, X., Tan, J., Li, J., Zhang, Z., Li, Q., Hu, X.: Refinemask: Towards high-quality instance segmentation with fine-grained features. In: Proceedings of the IEEE/CVF conference on computer vision and pattern recognition. pp. 6861--6869 (2021)

\bibitem{51}
Zhang, Z., Wang, H., Han, S., Dally, W.J.: Sparch: Efficient architecture for sparse matrix multiplication. In: 2020 IEEE International Symposium on High Performance Computer Architecture (HPCA). pp. 261--274. IEEE (2020)

\bibitem{18}
Zhong, Z., Liu, X., Jiang, J., Zhao, D., Ji, X.: Guided depth map super-resolution: A survey. ACM Computing Surveys  \textbf{55}(14s),  1--36 (2023)

\bibitem{streamvggt}
Zhuo, D., Zheng, W., Guo, J., Wu, Y., Zhou, J., Lu, J.: Streaming 4d visual geometry transformer. arXiv preprint arXiv:2507.11539  (2025)

\end{thebibliography}

\end{document}